\documentclass{article}

\usepackage{microtype}
\usepackage{graphicx}
\usepackage{subcaption}
\usepackage{booktabs} 
\usepackage{tabularx}   

\makeatletter
\def\Notice@String{}
\makeatother
\usepackage[preprint]{icml2026}
\usepackage{hyperref}

\usepackage[preprint]{icml2026}


\usepackage{amsmath}
\usepackage{amssymb}
\usepackage{mathtools}
\usepackage{amsthm}
\usepackage{xcolor}  
\usepackage{pifont}

\usepackage[capitalize,noabbrev]{cleveref}

\hypersetup{colorlinks=true, linkcolor=blue, urlcolor=blue}

\usepackage[most]{tcolorbox} 
\tcbuselibrary{skins,breakable}

\newtcolorbox{systemprompt}[1]{
    colback=white,          
    colframe=gray!60!black, 
    colbacktitle=gray!60!black, 
    coltitle=white,         
    fonttitle=\bfseries\small\sffamily, 
    enhanced,               
    attach boxed title to top left={yshift=-2mm, xshift=2mm}, 
    boxed title style={sharp corners=south, size=small}, 
    sharp corners,          
    drop shadow,            
    title={#1},             
    verbatim, 
    left=1mm, right=1mm, top=3mm, bottom=1mm 
}
\usepackage{graphicx} 

\newtcolorbox{problembox}[1]{
    enhanced,
    breakable,              
    colback=white,          
    colframe=blue!30,       
    sharp corners,          
    drop shadow={black!20}, 
    fontupper=\small,
    attach boxed title to top left={xshift=-5mm, yshift=-3mm},
    boxed title style={colback=blue!40, sharp corners, frame hidden},
    title={#1}
}

\newtcolorbox{answerbox}{
    enhanced,
    colback=blue!5!white,
    breakable,              
    colframe=blue!10,
    sharp corners,
    boxrule=0.5pt,
    fontupper=\small,
    top=3mm,
    break at=0pt,
    boxsep=1mm,
}
\newenvironment{answerblock}{
    \par
    \noindent
    \begin{minipage}{\linewidth}
    \vspace{3mm}
    \colorbox{blue!5!white}{
        \parbox{\dimexpr\linewidth-2\fboxsep}{
        \small
}{
        }
    }
    \vspace{3mm}
    \end{minipage}
    \par
}

\usepackage{listings}    
\theoremstyle{plain}

\theoremstyle{definition}

\theoremstyle{remark}

\usepackage[disable,textsize=tiny]{todonotes}

\icmltitlerunning{AIR: Adaptive Interleaved Reasoning with Code in MLLMs}

\begin{document}
\newcommand{\ourmethod}{AIR}
\newcommand{\cmark}{\ding{51}}%
\newcommand{\xmark}{\ding{55}}
\twocolumn[
  \icmltitle{\ourmethod: Adaptive Interleaved Reasoning with Code\\ in Multimodal Large Language Models}



  \icmlsetsymbol{equal}{*}

  \begin{icmlauthorlist}
    \icmlauthor{Cong Han}{ind}
    \icmlauthor{Xiaohan Lan}{ind}
    \icmlauthor{Haibo Qiu}{ind}
    \icmlauthor{Yujie Zhong}{ind}
  \end{icmlauthorlist}

  \icmlaffiliation{ind}{Independent Researcher}

  \icmlcorrespondingauthor{Cong Han}{hancong0911@163.com}

  \icmlkeywords{Machine Learning, ICML}

  \vskip 0.3in
]



\printAffiliationsAndNotice{}

\begin{abstract}
 
Following the paradigm shift initiated by OpenAI o3, 
interleaved reasoning with code to enhance 
multimodal large language models (MLLMs) has become a pivotal research frontier.
The existing literature focuses primarily on tool-use within vision-perception tasks.
However, such approaches typically rely on predefined heuristics for visual manipulation and are inherently incapable of addressing numerical computation problems due to their exclusive focus on visual operations.
This paper empowers MLLMs with adaptive interleaved reasoning capabilities through extended reinforcement learning training on code-augmented complex numerical computation tasks.
To this end, we propose a comprehensive three-component solution consisting of: a two-stage cold-start data construction pipeline, data filtering strategies for RL dataset curation, and an adaptive tool-invocation strategy leveraging a group-constrained reward function for interleaved reasoning trajectories.
Extensive experiments demonstrate that after Reinforcement Learning training with the group-constrained reward function, performance improves by an average of $6.1$ percentage points (pp) on evaluation benchmarks. 
Specifically, the accuracy for interleaved reasoning samples increases by $9.9$ pp, and the overall success rate of tool-use exceeds 95\%.
Our data and code are available at: \textbf{\url{https://github.com/CongHan0808/AIR.git}}.

\end{abstract}

\section{Introduction}
Since the release of DeepSeek-R1 \cite{deepseekai2025deepseekr1} and Kimi-K1.5 \cite{team2025kimi} in early 2025, the application of Reinforcement Learning from Verifiable Rewards (RLVR)~\cite{Wen2025ReinforcementLW, Shao2024DeepSeekMathPT} to enhance the capabilities of Large Language Models (LLMs) has garnered significant research interest. 
Agentic RL~\cite{Zhang2025TheLO} has emerged as a pivotal direction in this field. 
In particular, following the release of OpenAI o3 and o4-mini~\cite{openai2025thinking}, the paradigm of interleaving tool-use within the reasoning process has demonstrated a closer alignment with human cognitive patterns. 
In the domain of Multimodal Large Language Models (MLLMs), several open-source works \cite{Liu2023VisualIT,hong2025deepeyesv2, zheng2025deepeyes, zhang2025thyme} have explored the integration of interleaved tools\thinspace--such as rotation and zooming-- to effectively resolve complex visual perception and reasoning tasks.

In the context of MLLMs, tool invocation scenarios extend far beyond image-related operations; yet confining tool invocation to such visual-centric operations alone fails to meet the demands of multimodal reasoning, particularly for tackling problems involving complex numerical computation --for which the utilization of Python code as a computational auxiliary is essential to address intricate mathematical reasoning tasks.
We provide supplementary illustrative examples for this in \textbf{Appendix \ref{appendix:Example}}.
This capability requires the model to autonomously trigger Python scripts when encountering sophisticated calculations during textual reasoning, thereby simplifying the computational process and ensuring the precision of the final output. 
Such a paradigm closely mirrors human problem-solving behavior: when a solution cannot be derived through mental calculation alone, external calculators serve as indispensable tools. 
Despite its promise, the interleaved reasoning of this nature remains relatively under-explored in the open-source MLLMs community.

We explore adaptive tool invocation for complex numerical computation  tasks, proposing optimizations across data synthesis, filtering, and RL strategies. 
Given the difficulty of activating these functions in smaller open-source models, we introduce an essential cold-start framework. 
Specifically, we develop a two-step pipeline --incorporating textual CoT generation and code-augmented interleaved reasoning rewriting-- to streamline SFT data synthesis. 
We demonstrate that this decoupled architecture significantly reduces the complexity of generating high-quality training data. 
Unlike single-step synthesis, which often falters in reasoning depth and efficiency within multimodal scenarios, our approach leverages the inherent prompt-learning capabilities of LLMs and the availability of CoT data to ensure more effective and reliable data production.

To ensure the high fidelity of the Agentic RL process, we implement two distinct data curation strategies: Self-Sampled and Prior-Filtered Strategy. 
The former employs a multi-turn consensus mechanism, where the candidate's eligibility is determined by the Pass@k metric across multiple rollouts of the target model. 
Alternatively, the Prior-Filtered Strategy functions as a supplementary extrinsic quality gate, leveraging the robust evaluative capacity of high-capacity teacher models for rigorous verification.
This dual-track approach effectively mitigates noise in the training distribution, providing a stable foundation for complex policy optimization.

RL has been widely applied to the training of large language models (LLMs), as this paradigm can enhance the reasoning capabilities of models~\cite{Liu2025SuperRLRL,Xue2025SimpleTIRER,Singh2025AgenticRA}.
To activate the model's adaptive tool-invocation capability during the reinforcement learning phase, we have designed a group-constrained reward mechanism.
Through the optimization of this mechanism, the model can adaptively determine whether tool invocation is necessary and decide on the optimal timing for tool calls.
Furthermore, we made an unexpected finding that this mechanism also addresses the issue of unstable long-term training in Agentic RL~\cite{dong2025arpo}. 
We consistently observed this phenomenon in our experiments; notably, as the proportion of tool-use increased, this instability intensified significantly, in some cases leading to model collapse. 
Adaptive tool invocation facilitates stable long-term agentic RL training.

In summary, the primary contributions of this paper are threefold:
\begin{itemize}
    \item \textbf{A Two-Stage Data Construction Methodology:} We introduce a two-stage approach for synthesizing cold-start data that supports code-augmented interleaved reasoning.
    \item \textbf{Data Filtering Strategies:} We propose two data filtering methods, namely the Self-Sampled Strategy and Prior-Filtered Strategy, which effectively curates interleaved reasoning data.
    \item \textbf{An Adaptive Tool-Invocation Strategy:} We develop a GRPO with group constraints, which enables adaptive tool invocation and ensures the long-term stable training of Agentic RL models.
\end{itemize}

\section{Related Work}
\begin{figure*}[htbp]
    \centering
    \includegraphics[width=0.95\textwidth]{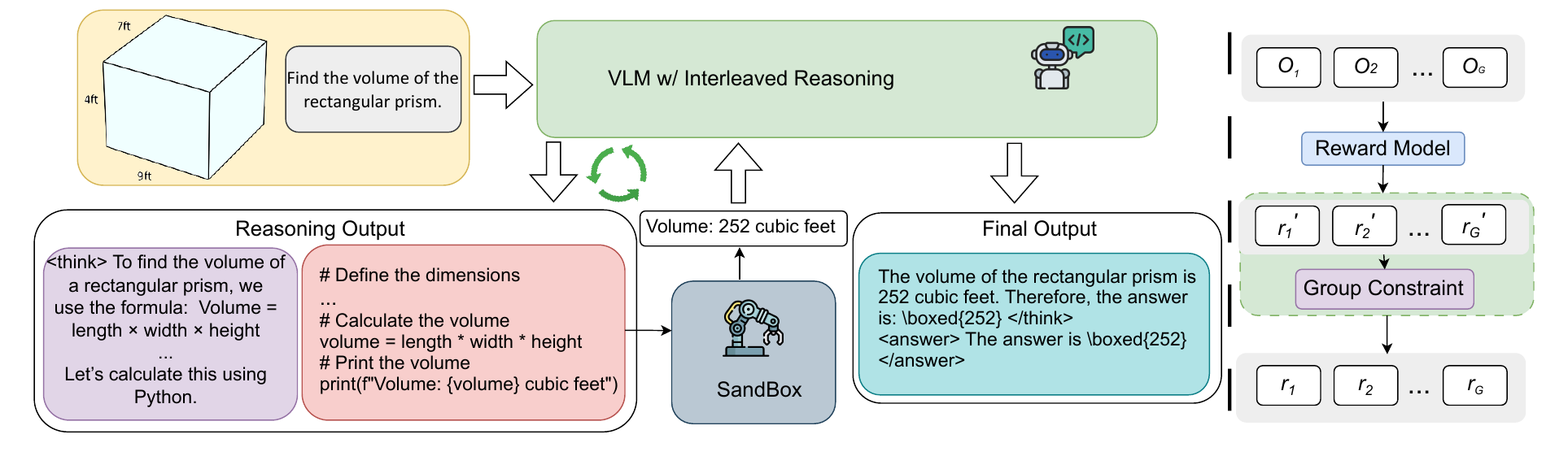}
    \caption{\textbf{Left:} Overview of \ourmethod. Upon receiving an image and a question, \ourmethod~ generates textual reasoning and optionally invokes a code tool. The code undergoes execution in a sandbox, and its output is fed back into the reasoning process. This cycle of text reasoning and code execution can be repeated multiple times until the final answer is produced. \textbf{Right:} Illustration of reward design with group constraints. After computing sequence rewards via the reward model, we introduce a group constraint on the proportion of interleaved reasoning sequences, then recalculate rewards and derive group relative advantages.}
    \label{fig:framework}
    \vspace{-2mm}
\end{figure*}

\paragraph{Multimodal Large Language Models (MLLMs).} 
Recent years have witnessed substantial advancements in Multimodal Large Language Models (MLLMs), which integrate visual and textual modalities to achieve sophisticated cross-modal understanding.
This rapidly evolving field is characterized by powerful closed-source systems such as the Gemini~\cite{Comanici2025Gemini2P} and GPT~\cite{openai2024learning} series, as well as highly capable open-source families, most notably the Qwen-VL~\cite{Qwen3-VL, Qwen2.5-VL}  and InternVL~\cite{wang2025internvl3_5, zhu2025internvl3} series. 
The success of these models stems from integrating high-capacity vision encoders with advanced Large Language Models (LLMs) through efficient cross-modal alignment modules. 
By using large-scale image-text pre-training followed by multimodal instruction tuning, these models have shown emergent abilities in complex reasoning, fine-grained visual recognition, and long-context multimodal understanding. 
The paradigm has shifted from basic image captioning to more interactive and open-ended tasks. This allows MLLMs~\cite{Liu2023VisualIT} to serve as general-purpose assistants capable of interpreting diverse visual inputs, such as natural scenes and document layouts, within a unified linguistic framework.

\textbf{Reasoning Model in MLLMs.}
Since the beginning of 2025, the paradigm of MLLMs has undergone a significant shift toward intrinsic reasoning, exemplified by the release of DeepSeek-R1 and its subsequent multimodal extensions. 
Unlike traditional models that rely on direct visual-to-text mapping, these next-generation MLLMs incorporate Large-scale Reinforcement Learning (RL)—specifically Reinforcement Learning from Verifiable Rewards (RLVR)—to stimulate system-2 'slow thinking' capabilities. 
This approach~\cite{qiu2025metis, Yue2025MiMoVLTR,Liu2025VisualRFTVR} has yielded remarkable breakthroughs in tasks requiring multi-step logical deduction, such as complex mathematical reasoning, fine-grained visual localization, and hierarchical OCR parsing. 
However, a critical bottleneck persists: most current reasoning-based MLLMs are optimized for static linguistic justifications within closed-loop environments. 
Their ability to autonomously invoke external tools or execute multi-stage actions remains constrained by the lack of executable feedback loops and the misalignment between internal reasoning traces and external API requirements. 
This gap highlights a pressing need for algorithmic innovations that can bridge the divide between multimodal deliberation and agentic execution.

\paragraph{Tool Use in MLLMs.}
Building upon the advancements in general-purpose Large Language Models (LLMs) and their subsequent multimodal integration, recent progress has centered on augmenting MLLMs with external tool-use mechanisms~\cite{feng2025retoolreinforcementlearningstrategic,Li2025Searcho1AS,Yang2025VisionThinkSA}. 
These models have demonstrated impressive performance in both closed-source and open-source ecosystems. 
A pivotal moment occurred, particularly since the release of OpenAI o3 in early 2025, which catalyzed research on tool-augmented visual reasoning, making it increasingly sophisticated. 
Representative works include the DeepEyes series~\cite{hong2025deepeyesv2,zheng2025deepeyes} and VTool-R1~\cite{wu2025vtoolr1vlmslearnthink} , which enable models to analyze visual information by dynamically invoking tools for operations such as rotation, cropping, and zooming before performing secondary reasoning. 
These methods have yielded significant improvements in fine-grained perception and detailed chart-based reasoning. 
Nevertheless, a conspicuous void exists in the current literature. Existing research primarily focuses on enhancing the model’s perceptual capabilities (System 1 cognition)~\cite{watson2011d}. 
Exploration of computation-intensive scenarios, which are common in complex reasoning, remains insufficient. 
To address this critical gap, the core contribution of this work is to empower MLLMs with the robust ability to invoke and orchestrate specialized tools specifically for such computation-heavy tasks, thereby advancing System 2~\cite{watson2011d}reasoning in MLLMs.
\section{Methods}
In this section, we introduce \ourmethod, a framework designed to resolve complex mathematical problems through an interleaved reasoning process powered by adaptive tool invocation. 
We first provide an overview of the proposed architecture in Section \ref{method-overview}, as illustrated in Figure \ref{fig:framework}. 
Subsequently, Section \ref{method-sft} details the construction pipeline of the cold-start dataset. 
Finally, we describe the reinforcement learning (RL) phase in Section \ref{method-rl}, which is specifically designed to bolster the model’s capability for adaptive interleaved reasoning.
\vspace{-2mm}
\subsection{Overview}
\label{method-overview}

Our approach consists of two training stages: cold-start Supervised Fine-Tuning (SFT) and interleaved reasoning Reinforcement Learning. 
To support these stages, we designed two dedicated data pipelines: an SFT data construction pipeline and an RL data filtering pipeline.

The SFT data construction pipeline consists of a two-step generation process --textual CoT generation by an MLLM and interleaved reasoning data production by an LLM-- followed by a final synthesis stage. 
In this stage, the cold-start dataset is curated by rigorously verifying code execution integrity and result accuracy.

The RL data filtering pipeline bifurcates into two streams: Self-Sampled Data, derived from SFT model inferences via tool-call success rates, and Prior-Filtered Data, identified by a robust prior model. 
During RL training, a group-constrained reward is implemented to foster adaptive and precise tool invocation.

As illustrated in Figure \ref{fig:framework}, the overview of \ourmethod~ comprises two main components: a Multi-modal Large Language Model (MLLM) capable of interleaved text/code reasoning and a sandbox environment. 
The MLLM receives the user-provided image and question, performs textual reasoning, and adaptively determines whether to generate code. 
The code is then executed within the sandbox, with the results returned to the MLLM. 
This process iterates until the MLLM produces the final output.

\begin{figure*}[htbp]
    \centering
    \includegraphics[width=0.85\textwidth]{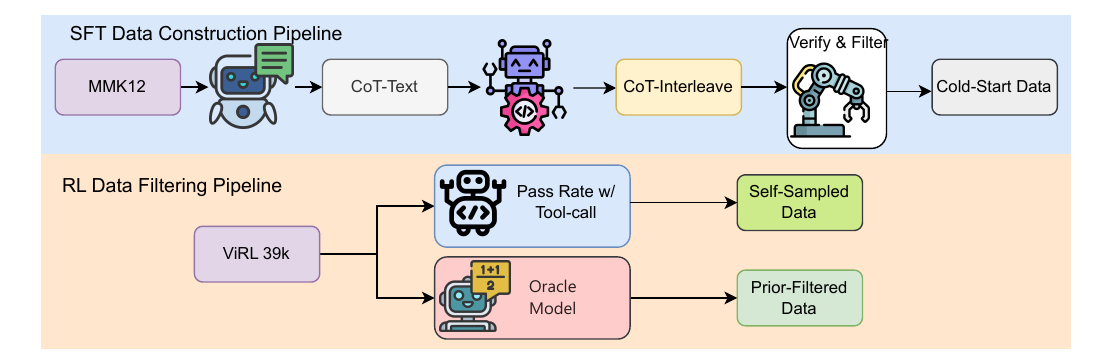}
    \caption{\textbf{Top:} SFT data construction pipeline; \textbf{Bottom:} RL data filtering method. \textbf{SFT Data Construction Pipeline}: For MMK12~\cite{meng2025mmeureka} data, powerful MLLMs and LLMs generate text-only and text-code interleaved reasoning data (CoT-Text and CoT-Interleave). Cold-start Data is then produced after verifying code execution and result accuracy in a sandbox. \textbf{RL Data Pipeline}: For the ViRL39k~\cite{Wang2025VLRethinkerIS} dataset, two data batches (self-sampled and prior-filtered data) are filtered based on the student model's tool call rate across multiple roll-outs and labels from an oracle model.}
    \label{fig:data_flow}
    \vspace{-2mm}
\end{figure*}
\subsection{Constructing Cold-Start Data via Interleaved CoT Synthesis} 
\label{method-sft}
Given that existing MLLMs typically struggle with interleaved reasoning in mathematically intensive tasks, a cold-start SFT stage is indispensable. 
We streamline the creation of such multimodal data through a two-stage approach: generating initial textual CoT via an MLLM and subsequently restructuring it into interleaved reasoning traces using an LLM. 
This pipeline, illustrated in Figure~\ref{fig:data_flow}, adopts a cascaded construction paradigm and undergoes rigorous validation to finalize the dataset.
This strategy effectively lowers the technical barriers to constructing complex reasoning trajectories.

\textbf{Text reasoning data} was synthesized using Gemini 2.5 Pro~\cite{Comanici2025Gemini2P}, a closed-source multimodal large language model (MLLM).  
By feeding images and questions into the reasoning-capable MLLM, a detailed reasoning path was generated for each sample. 
To ensure the reliability of the synthesized data, a consistency filter was subsequently applied, retaining only the entries where the model’s final predictions were consistent with the original ground-truth answers.
\textbf{Interleaved reasoning data} refers to a data form where text and code tools appear alternately: the outputs generated by executing code in a sandbox environment are fed back into the model to support reasoning in the subsequent round, which ultimately yields the final result.
This data form is derived by refining the original text reasoning data.
Specifically, we designed a task-specific system prompt ( see Appendix~\ref{appendix:system_prompt} for details) that is tailored to interleaved reasoning scenarios, which guides the model to perform high-quality data synthesis.
We then fed the text reasoning model into Gemini and replaced its numerical calculation components with executable Python scripts. 
The mathematical formulation of interleaved reasoning data is:
\begin{equation}
    \label{eqa:o_interleave}
    s = \{ (I, Q) | (T_1, C_1, A_1),... (T_{i}, C_{i}, A_{i}), (T, Ans) \}
\end{equation}
where $I$ denotes the input image, and $Q$ represents the input question; $T$ is text reasoning, $C$ means Python code snippet, 
and $A$ is the execution result; $Ans$ is the final concise answer.

\textbf{Verify and Filter,} the final step in data construction, ensures the correctness and usability of the resulting dataset.
To this end, we adopted the rationality of the reasoning process and the accuracy of the results as core criteria to verify and filter the interleaved reasoning data.
In reasoning process verification, we extracted the code segments and their corresponding results from each interleaved data entry, which are enclosed in the \texttt{<code></code>} and \texttt{<interpreter></interpreter>} tags, respectively.
The extracted code was executed in a sandbox environment, and we verified the consistency between the execution output and the reasoning result.
For result correctness, we extracted the answers from the \texttt{<answer></answer>} tags of the synthetic data and compared them against the ground truth to confirm their accuracy.
In summary, the cold-start dataset must ensure that the code in each data sample can run correctly and that the corresponding answer is accurate.

\subsection{Reinforcement Learning for Adaptive Interleaved Reasoning}
\vspace{1mm}
\label{method-rl}
Adaptive interleaved reasoning requires the model to autonomously make two key decisions during the reasoning process: first, whether to adopt the interleaved reasoning paradigm itself; and second, when to perform textual reasoning or invoke tools within the chosen paradigm. This dual-autonomy capability can be effectively enhanced through reinforcement learning.
This chapter elaborates on the reinforcement learning methods for improving the model’s adaptive interleaved reasoning performance.
To boost this capability, we provide a detailed introduction to training data filtering and reward strategy design.
\subsubsection{Data Filtering Strategies for Reinforcement Learning}
\vspace{1mm}
The quality of data serves as a core prerequisite for the effectiveness of RL in enhancing adaptive interleaved reasoning. 
Poorly curated data not only introduces noise into the training process but also leads to suboptimal policy convergence and degraded tool-invocation decision-making performance of the model. 
During the training of interleaved reasoning tasks, inappropriate data may fail to effectively activate the model's tool-invocation capability or lead to excessive tool invocation.
To address this issue, this subsection presents two targeted data filtering strategies, \textbf{Self-Sampled} and \textbf{Prior-Filtered}, tailored for RL-driven autonomous interleaved reasoning tasks. 

\textbf{Self-Sampled Strategy} leverages the inherent capabilities of the model for data filtering, with the specific goal of identifying whether a given data sample is suitable for resolution via tool invocation.
Specifically, for an individual data prompt denoted as $p_i$, we first guide the model to perform $N$ roll-outs under two distinct reasoning paradigms --textual reasoning and interleaved reasoning-- by means of system prompts.
We then calculate the success rate of each paradigm, denoted respectively as $S_{Ti}$ for textual reasoning and $S_{Ii}$ for interleaved reasoning.
Subsequently, we determine whether to retain the data sample $p_i$ by verifying the decision condition $I(p_i) = (S_{Ii}-S_{Ti}-\delta)>0$.
Here, $\delta$ serves as a tunable threshold (ranging from $0$ to $1$) that balances the stringency of the sample retention criterion.
Rooted in the model’s self-capabilities for data filtering, this method exhibits strong effectiveness under the reinforcement learning training paradigm widely adopted by modern LLMs.
Beyond standalone use, it can be integrated with other rejection sampling methods to further enhance the quality of the resulting training data.
Furthermore, for student models, this strategy supports iterative re-filtering with the updated model following each round of training, thereby constructing an efficient data flywheel that improves data utilization efficiency and drives consistent gains in model performance.

\textbf{Prior-Filtered Strategy} focuses on labeling each data sample in the dataset to indicate whether it is suitable for resolution via tool invocation—an essential process that hinges on prior knowledge to render judgments on individual samples.
Among all available sources of prior knowledge, human experts possess the most robust expertise; yet, manual annotation is plagued by prohibitively high costs, making large-scale implementation impractical.
Notably, humans have developed state-of-the-art large language models (LLMs) such as GPT~\cite{openai2025gpt}, Gemini~\cite{Comanici2025Gemini2P}, and Claude~\cite{anthropic2025claude}, which have already demonstrated exceptional reliability in code-centric tool invocation tasks.
Building on this observation, we posit that these advanced LLMs also encapsulate abundant prior knowledge, enabling them to accurately judge whether a given data sample necessitates tool invocation for effective resolution.
To operationalize this insight, we design rigorous prompts and evaluation criteria, which are jointly fed into the LLM alongside the target data sample $p_i$.
Leveraging in-context learning and the model’s inherent capabilities, the proposed framework generates precise labels to determine the retention of the data sample $p_i$.
\subsubsection{Roll-out with Interleaved Reasoning}
\vspace{1mm}
During the reinforcement learning training for interleaved reasoning, it is necessary to input the code segment generated in the roll-out process into a sandbox environment for execution and validation.
We require the sandbox environment to output the execution result in text format each time, which is the key to interacting with the environment and conducting continuous reasoning.
Subsequently, we embed the text output by the sandbox within the \texttt{<interpreter></interpreter>} tags, concatenate it with the textual reasoning output, and feed the combined sequence into the model for the next round of reasoning until the final result is generated.
We denote the model input for the next round as $PR_{i+1}= \text{Concat}(PR_{i},(T_{i},C_{i},A_i))$, where $PR_i$ represents the input for each reasoning round, and $PR_0=(I,q)$ is the original input consisting of an image $I$ and a question $q$, with the initial states set as $T_0=\text{None}$, $C_0=\text{None}$, and $A_0=\text{None}$.

\subsubsection{Reinforcement Learning Algorithm}
\vspace{1mm}
To tackle the challenges of adaptive interleaved reasoning for large language models (LLMs), this section leverages the Group Relative Policy Optimization (GRPO)~\cite{Shao2024DeepSeekMathPT} algorithm—a widely adopted paradigm for LLM-based reinforcement learning tasks.
A key enhancement is made to the reward function by incorporating intra-group constraints shown in the right panel of Figure~\ref{fig:framework}, which enables the model to achieve effective adaptive interleaved reasoning. 
Additionally, several improved variants of the GRPO algorithm, such as DAPO~\cite{Yu2025DAPOAO} and GSPO~\cite{Zheng2025GroupSP}, can also incorporate this method for autonomous tool-invocation training in Agentic RL.
The subsequent content presents a detailed breakdown of the training algorithm and the tailored design of the reward function.

\textbf{GRPO} is a reinforcement learning algorithm designed to LLMs, while achieving a significant reduction in training costs compared to traditional methods such as PPO (Proximal Policy Optimization)~\cite{Schulman2017ProximalPO}.
For each input prompt $Q$ composed of an image $I$ and a text question $q$, the model generates a group of $G$ outputs $\{o_1, o_2, \ldots, o_G\}$ using the current old policy $\pi_{\theta_{\text{old}}}$. Each trajectory $o_i$ includes text-based reasoning content, and may also incorporate code and its execution results; details are presented in Equation \ref{eqa:o_interleave}.

Subsequently, a reward model is employed to score each trajectory, yielding a set of $G$ rewards denoted as $\textbf{r} = \{r_1, r_2, \ldots, r_G\}$. These rewards are then normalized by subtracting the group average and dividing by the group standard deviation.

Outcome supervision assigns the normalized reward to the end of each output $o_i$, and sets the advantage $A_{i,t}$ of all tokens within the output equal to this normalized reward, which is formally expressed as:
\begin{equation}
\label{eqa:gae}
    A_{i,t} = \tilde{r_i} = \frac{r_i-mean(\textbf{r})}{std(\textbf{r})}.
\end{equation}

Then, the policy model is optimized by maximizing the following objective:
\begin{equation}
\begin{split}
    J_{GRPO}(\theta) = \mathbb{E}_{PR\sim D,\,\{Q_i\}_{i=1}^{G}\sim \pi_{old}} \bigl[
    L_{clip}(o_{i,t}) - {} \\
    \beta D_{KL}\bigl(\pi_{\theta}\;\big\|\;\pi_{ref}\bigr)
    \bigr]
\end{split}
\end{equation}
where $L_{clip}(o_{i,t})$ is the clipped surrogated loss, defined as:
\begin{equation}
\begin{split}
L_{\text{clip}}(o_{i,t}) = \min\bigg[ & \frac{\pi_{\theta}}{\pi_{\theta_{\text{old}}}} A_{i,t}, \\
& \text{clip}\bigg(\frac{\pi_{\theta}}{\pi_{\theta_{\text{old}}}}, 1-\epsilon, 1+\epsilon\bigg) A_{i,t} \bigg]
\end{split}
\end{equation}
where $\epsilon$ and $\beta$ are hyper-parameters, $A_{i,t}$ is the advantage calculated by Equation \ref{eqa:gae}.

The \textbf{Reward Function} is used to assign a reward to each output trajectory, serving as the source of training signals and determining the direction of RL optimization.
We adopt a rule-based approach to provide precise feedback for each data sample.
Following DeepSeek-R1~\cite{deepseekai2025deepseekr1}, we employ a hybrid reward structure consisting of a format reward, a correctness reward, and a code-integrated reward. 
The code-integrated reward is unique to interleaved reasoning and also constitutes a key component of group constraints.
We now provide a detailed breakdown of these reward functions.

\textbf{Format Reward} requires the trajectories output by the model to strictly comply with specific format requirements, and this reward can be denoted as $\mathbf{R_{f}}$.
Specifically, the reasoning process output by the model should be enclosed within the designated tag \texttt{<think></think>}; a concise description of the results should be placed in \texttt{<answer></answer>}; and numerical values or definitive outcomes need to be encapsulated in \texttt{\textbackslash box{}} and conform to LaTeX syntax.
During interleaved reasoning, the code script blocks output by the model must be wrapped in \texttt{<code></code>}, and the direct results corresponding to the code blocks should be placed immediately after the respective code blocks and enclosed within \texttt{<interpreter></interpreter>}.
If the format meets the requirements, $R_f=0.5$; otherwise, $R_f=0.0$.

\textbf{Accuracy Reward} assesses the correctness of model responses. 
Specifically, upon completion of the model roll-out, the system first extracts result information from the generated output— which is located within the tag \texttt{<answer></answer>} —  and then validates it against the ground-truth (gt) values. 
A reward of $\mathbf{R_{acc}}$ is subsequently assigned to the corresponding output trajectory based on these validation results. 
If the result is correct, $R_{acc}=1.0$; otherwise, $R_{acc}=0.0$.

\textbf{Code-Integrated Reward with Group Constraint} is specially designed for the code snippets in interleaved reasoning.
For the code integrate reward component, we define an initial reward value denoted as $\mathbf{R_{raw}}$, where $R_{raw}$ is set to 0.5.
To account for the success rate of tool execution, each trajectory is assigned a baseline reward: \(\mathbf{R_{base}}=R_{raw}\times \mu \), where $\mu = \tfrac{N_{succ}}{N_{total}}$ represents the ratio of successful code invocations $N_{succ}$ to the total number of attempts $N_{total}$.
Furthermore, we incorporate the influence of intra-group tool-calling distribution, specifically the proportion of sequences utilizing tools relative to those without. 
We differentiate this proportion based on the correctness of the final answer, denoted as \(P_{r}\) (correct) and \(P_{w}\) (incorrect). 
The decoupled adjustment of the upper and lower bounds for \(P_{r}\) and \(P_{w}\) is the pivotal factor in achieving autonomous tool invocation. 
Specifically, a lower bound \(P_{min}=0.0\) imposes no constraints on sparse tool usage, while an upper bound $P_{max}=1.0$ indicates that every trajectory involves tool calls.
In summary, the reward function for tool invocation is formulated as follows:
\begin{equation}
  R_{\text{code}} =
    \begin{cases}
    R_{\text{base}}, \quad  \text{if} \  R_{\text{acc}} > 0 \ \text{and} \  P_{\text{rmin}} < P_r < P_{\text{rmax}} \\
                      
    R_{\text{base}}, \quad \text{if} \  R_{\text{acc}} = 0 \ \text{and} \  P_{\text{wmin}} < P_w < P_{\text{wmax}} \\
    0,               \quad \text{others}
    \end{cases}
\end{equation}

In summary, the reward design for interleaved reasoning adopts a hybrid reward scheme that incorporates the format reward, accuracy reward, and tool integration reward. The final reward can be expressed as:
\begin{equation}
R = R_{f} + R_{acc} + R_{code}.
\end{equation}
Unlike the conventional GRPO, which only considers individual data samples, we introduce intra-group constraints to optimize reward calculation, as illustrated in the right panel Figure~\ref{fig:framework}. 
This design aims to enhance the model’s decision-making capability for interleaved reasoning and enable adaptive tool invocation.

\section{Experiment}
\begin{table*}[t]
\begin{center}
  \caption{Results on Multimodal Reasoning Benchmarks. $*$ denotes the results of self-evaluation. 
  The $Tool$ column indicates whether the model is equipped with interleaved reasoning. }  %
  \label{tab-main_result}
  \vspace{2mm}
  \centering  %
  \begin{small}
    \begin{tabular}{lcllllll}  %
      \toprule
      Model & Tool  & MathVista & MathVision & MathVerse & DynaMath & WeMath & LogicVista  \\
      \midrule
      Qwen2.5VL-7B\textsuperscript{*}~\cite{Qwen2.5-VL} & \xmark & 64.1 & 24.6 & 40.9 & 53.5 & 30.7 & 44.3  \\  %
      Qwen2.5VL-7B w/ GRPO & \xmark & 71.8 & 27.8 & 44.7 & 55.1 & 40.5 & 47.2 \\
      MM-Eureka~\cite{meng2025mmeureka} &\xmark&  72.6 & 28.1 & - &- & 21.8 & -\\       %
      VL-Rethinker~\cite{Wang2025VLRethinkerIS}&\xmark &  \textbf{73.7} & 28.4& - &- & 36.3 & 45.9 \\       %
      Thyme~\cite{zhang2025thyme} &\cmark&  70.0 & 27.6 & - &-  &39.3 & - \\       %
      DeepEyes~\cite{zheng2025deepeyes} &\cmark & 70.1 &26.6 & 47.3 & 55.0 & 38.9 &  47.7 \\
      \textbf{\ourmethod}-SFT(ours)&\cmark &  67.8& 27.9 & 43.5& 48.3& \textbf{43.9} & 43.2 \\
      
      \textbf{\ourmethod}(ours)\textsubscript{\small{ $\Delta$ vs Qwen2.5VL-7B}} & \cmark  &  72.2\textsubscript{\small{+8.1}}  & \textbf{28.4} \textsubscript{\small{+3.8}} & \textbf{48.6}\textsubscript{\small{+7.7}} & \textbf{57.5}\textsubscript{\small{+4.0}} & 40.6\textsubscript{\small{+9.9}} & \textbf{47.7}\textsubscript{\small{+3.4}} \\ 
      \bottomrule
    \end{tabular}
  \end{small}
\end{center}
\end{table*}  %
\vspace{2mm}
\begin{table*}[!htbp]
\begin{center}
  \caption{Comparison between interleaved reasoning ratio, code execution accuracy and Evaluation Set Metrics. 
    \textbf{Model-low/mid/high} denote the levels of interleaved reasoning ratio of the model. 
    For each specific dataset, the two numerical values respectively represent the proportion of interleaved reasoning data and the execution accuracy of tools.
    \textbf{Avg.Acc.} indicates the average accuracy of the corresponding model on all data within each dataset.
     All numerical values mentioned above are expressed as percentages (\%).
  }  %
  \label{tab-toolcall-ratio}
  \vspace{2mm}
  \centering  %
  \begin{small}
    \begin{tabular}{cccccccc}  %
      \toprule
      Model  & MathVista & MathVision & MathVerse & DynaMath & WeMath & LogicVista & Avg. Acc.  \\
      \midrule
      model-low & 8.6 / 97.7 & 4.1 / 85.9 & 7.5 / 96.6 & 12.5 / 95.3 & 3.7 / 99.1 & 3.4 / 93.3 & 48.4 \\
      model-mid &15.4 / 98.1 & 7.8 / 86.9  &13.7 / 98.2 & 19.1 / 95.0 & 7.5 / 99.0 & 6.5 / 100 & 48.8 \\
      model-high &79.6 / 95.3 & 56.7 / 91.9 & 62.2 / 96.1 & 76.7 / 91.8 & 70.0 / 98.2 & 65.3 / 92.4 & 49.0 \\
      \bottomrule
    \end{tabular}
  \end{small}
\end{center}
\end{table*}

\subsection{Experimental Setup}
\vspace{2mm}
To verify the validity of the data and the feasibility of the method, we conducted experimental validation based on Qwen2.5 VL\cite{Bai2025Qwen25VLTR}.
As mentioned above, we utilized cold-start data in the SFT phase to activate the interleaved reasoning capability, and then employed filtered data and a specially designed reward function in the RL phase to enhance this capability.

Specifically, the cold-start data was constructed based on MMK12~\cite{meng2025mmeureka}, resulting in 1.5k high-quality interleaved reasoning samples. With the help of LlamaFactory~\cite{zheng2024llamafactory}, we performed lightweight training for 1 epoch on this dataset and developed the \ourmethod-SFT model.
In the RL training phase, the training data were filtered from ViRL39K~\cite{Wang2025VLRethinkerIS}. 
The \textbf{Self-Sampled} strategy filters approximately 6k data samples, while the \textbf{Prior-Filtered} strategy identifies 7k data samples.
We further modified MMEureka~\cite{meng2025mmeureka} by adding a sandbox for running Python scripts during the roll-out process.
Regarding the evaluation tools, we implemented targeted improvements to VLMEvalKit~\cite{duan2024vlmevalkit}.
While maintaining compatibility with its original evaluation methods, we integrated support for interactions with the sandbox environment to accommodate interleaved reasoning scenarios.
We selected mathematical reasoning benchmarks for evaluation to fair comparison with other methods; specifically, we use the datasets: MathVista~\cite{Lu2023MathVistaEM}, MathVision~\cite{wang2024measuring}, MathVerse~\cite{zhang2024mathverse}, DynaMath~\cite{Zou2024DynaMathAD}, WeMath~\cite{Qiao2024WeMathDY}, and LogicVista~\cite{Xiao2024LogicVistaML}.
In the evaluation, Qwen-Max~\cite{qwen3max} was employed as the judge model to score the output results.

\vspace{-4mm}
\subsection{Main Results}
\vspace{-2mm}

We conducted the training in two sequential phases: first, we leveraged cold-start data to complete SFT; then, in the RL phase, we optimized the reward design by incorporating the group constraint mechanism based on RL-specific data, and finally yielded the \ourmethod~model.
We present the model performance on six benchmarks in Table \ref{tab-main_result}, 
which demonstrates that the model already yields performance improvements after SFT training, with significant further gains achieved following subsequent RL training.
Specifically, in the full-scale evaluation, \ourmethod~ outperforms Qwen2.5VL-7B by \textbf{8.1 percentage points (pp)} on MathVista, \textbf{3.8 pp} on MathVision, \textbf{7.7 pp} on MathVerse, \textbf{4.0 pp} on DynaMath, \textbf{9.9 pp} on WeMath, and \textbf{3.4 pp} on LogicVista.
These results confirm that interleaved reasoning enhanced by code integration can markedly strengthen the model’s mathematical-domain competence; a thorough analysis of the benefits derived from interleaved reasoning will be provided in subsequent ablation experiments.

\begin{figure}[!t] 
    \centering         
    \includegraphics[width=0.8\columnwidth]{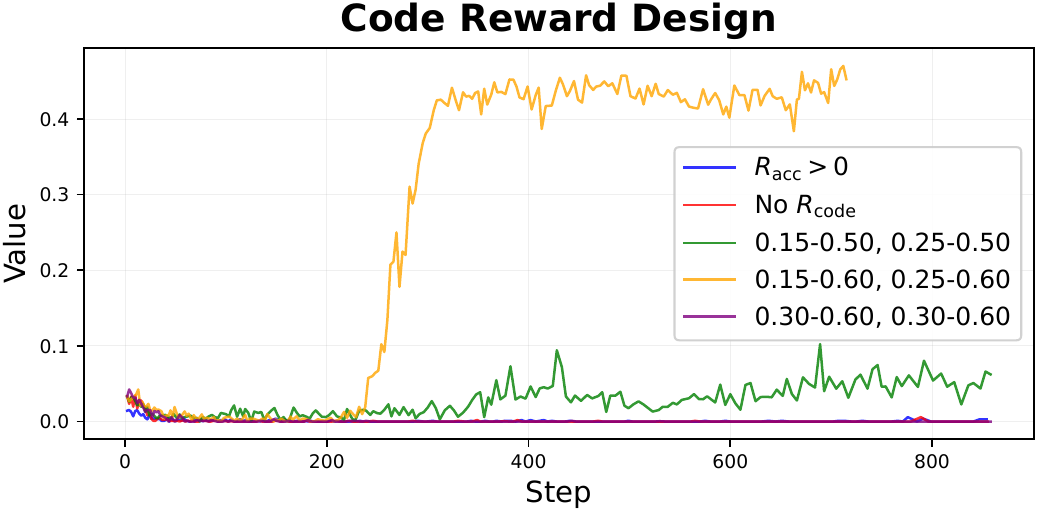} 
    \caption{ Impact of Reward Function Design on Code Rewards. $\textcolor{blue}{R_{\text{acc}} > 0}$ indicates that a correct result is the sole necessary condition. $\textcolor{red}{No \ R_{code}}$ means that the code reward does not provide supervision signals for RL, with only $R_{\text{code}}$ recorded. For other numerical parameters, the first and second parts of each row denote the lower and upper bounds of $P_{\text{r}}$ and $P_{\text{w}}$, respectively. } 
    \label{fig:cr_thres}
\end{figure}

\subsection{Ablation Study and Key Insight}
The \textbf{impact of tool-call ratios on evaluation results} is presented in Table 2.
Across various tool-call ratios, the success rate of tool usage consistently remains above \(95\%\). This indicates that the model has been effectively empowered to utilize tools correctly, which serves as the foundation for achieving interleaved reasoning. Furthermore, while evaluation metrics show a slight upward trend as the tool-call ratio increases, a substantial rise from \(10\%\) to approximately \(70\%\) yields only a \(0.2\,\text{pp}\) improvement, representing a poor trade-off in terms of efficiency. Consequently, it is crucial to equip the model with the capability to adaptively determine whether a tool call is necessary.

We conduct an ablation study to investigate the \textbf{effects of RL training data strategies}. The results are presented in Table~\ref{tab-ablation_data}.
We evaluated the impact of \textbf{self-sampled} and \textbf{prior-filtered} strategies on model performance individually, and further validated their performance by merging both datasets. 
Based on the average metrics of the evaluation set, the self-sampled strategy shows a slight advantage, while the prior-filtered strategy also demonstrates effectiveness in data selection. 
The self-sampled strategy is a customized approach that relies on the model's intrinsic capabilities for filtering, whereas the prior-filtered strategy provides a more objective selection with more generalizable results.

A quantitative comparison of \textbf{different paradigms in the interleaved reasoning scenario} is summarized in Table~\ref{tab-interreasoning_result}.
Given the insufficient interleaved reasoning samples in WeMath and LogicVista, we opt not to include these two benchmarks in the ablation analysis.
We observed that the average accuracy of using interleaved reasoning samples increased from $42.6$ to $52.5$, representing an improvement of $9.9$ pp. 
In contrast, the full benchmark dataset only achieved an improvement of $5.9$ pp (from $45.8$ to 5$1.7$). 
These results demonstrate that interleaved reasoning offers distinct advantages.

This section analyzes the \textbf{influence of interleaved reasoning on training stability.}
Figure~\ref{fig-compare_codereward} shows the variation curves of code reward and sequence entropy during the training process under three scenarios: no invocation, moderate invocation, and excessive invocation.
A Code Reward of 0 indicates that no interleaved reasoning occurs at all, corresponding to the scenario where the entropy of the output sequence remains within a reasonable range throughout the training process. 
When the Code Reward increases to around 0.1, abnormal peaks in the entropy of the output sequence emerge sporadically, which can return to normal quickly, ensuring overall training stability.
However, when the Code Reward approaches 0.5—meaning that interleaved reasoning is involved in nearly all output sequences—the entropy of the output sequence rises continuously. After approximately 750 steps, the training process collapses.
Thus, in the RL training for interleaved reasoning, prolonged excessive tool invocation will cause a continuous increase in the entropy of the output sequence and eventually lead to training collapse.

\begin{table}[!t]
  \centering      
  \small       
  \caption{Ablation study on RL training data. We validate the standalone efficacy of the \textbf{Self-Sampled} and \textbf{Prior-Filtered} strategies, along with the efficacy of the \textbf{Merge} method that fuses the two data sources.
  \textbf{M.Vista/M.Vision/M.Verse/Dyna} denote MathVista, MathVision, MathVerse, and DynaMath, respectively.}

  \label{tab-ablation_data}
  
  \begin{tabularx}{0.95\columnwidth}{c@{\hspace{0.5em}}c@{\hspace{0.5em}}c@{\hspace{0.5em}}c@{\hspace{0.5em}}c@{\hspace{0.5em}}c}
      \toprule
      Data & Avg. & M.Vista & M.Vision & M.Verse  & Dyna   \\
      \midrule
      Self-Sampled& 51.3 & 72.0 & 27.9 & 49.4 & 55.9 \\
      Prior-Filtered & 50.8 & 73.1 & 29.7 & 45.3 & 55.2 \\
      Merge &50.5  & 72.6 & 28.8  & 44.5 & 56.2 \\
      \bottomrule
  \end{tabularx}
\end{table}
\begin{table}[!t]     
  \centering             
  \small  
  \caption{Comparison Under the Interleaved Reasoning Scenario.
  \textbf{Q.VL-7B} means Qwen2.5VL-7B;
    \textbf{Avg.} = average accuracy on full benchmark; \textbf{A.I.R.} = average accuracy on interleaved reasoning subset.}
  \label{tab-interreasoning_result}
  
  \setlength{\tabcolsep}{3pt} 
  \begin{tabular}{ccccccc} 
    \toprule
    Model         & Avg.  & A.I.R & M.Vista & M.Vision & M.Verse & Dyna \\
    \midrule
    Q.VL-7B & 45.8    & 42.6   & 63.6     & 17.8  & 35.2  & 53.6     \\
    \ourmethod-SFT & 46.9   & 43.2  & 63.4    & 20.4    & 42.6    & 46.2     \\
    \ourmethod   & 51.7& 52.5 & 67.5  & 25.7    & 56.5     & 60.0       \\
    \bottomrule
  \end{tabular}
  \normalsize
\end{table}

How \textbf{reward function design affects code rewards} is illustrated in Figure~\ref{fig:cr_thres}.
The y-axis denotes the code reward in RL training, as defined in Figure~\ref{fig:cr_thres}. Three code reward function designs are compared: No $R_{code}$, \(R_{acc}>0\), and with group constraints.
$No \ R_{code}$ indicates that the code reward does not affect advantage calculation and thus provides no supervision signal for the training process. In this setting, the code reward is fixed at 0, which denotes the absence of an interleaved reasoning capability.
\textbf{$R_{acc}>0$} means that the code reward requires results to be correct. Under such strict constraints, interleaved reasoning rarely occurs.
Therefore, we argue that it is essential to decouple the code reward from the accuracy reward.
This design aligns with the intuition that task success is multi-faceted—influenced by problem difficulty and the overall coherence of the reasoning chain—of which tool usage is only one component. 

We evaluate the \textbf{impact of group-constrained reward thresholds} in this ablation experiment.
To compute the group-level tool reward, we analyzed the tool invocation frequencies separately for instances with correct and incorrect final answers, and we demonstrate the impact of different thresholds on the activation of tool invocation capability in Figure \ref{fig:cr_thres}.
Specifically, for instances where the final answer is correct, we assume the tool usage is more likely to be rational and thus set a lower threshold to encourage such behavior. 
Conversely, for incorrect instances, a higher threshold is applied to impose a stricter penalty, thereby steering the model toward more effective problem-solving rather than superficial tool usage.

Our experiments reveal that the lower-bound threshold significantly dictates the difficulty of activating tool-calling behaviors. 
A higher lower bound necessitates a greater number of tool calls before a reward is triggered, thereby increasing the activation barrier. 
Meanwhile, the upper-bound threshold influences the learning rate and convergence speed. 
While a higher upper bound can accelerate the acquisition of tool-calling skills, it also poses a risk of over-invocation, where the model may resort to redundant or unnecessary tool usage.
\subsection{Qualitative Analysis}
In Appendix C, we present examples of code-augmented interleaved reasoning, which are sourced from the aforementioned benchmark datasets.
It can be observed that the model is capable of autonomously determining the timing of tool invocation; after interacting with the sandbox environment, it proceeds with reasoning and ultimately derives the correct answer.
We also find that the model may interact with the sandbox environment multiple times. Furthermore, when errors occur during code execution, the model can correct them independently, which is highly consistent with the human reasoning process.

\section{Conclusion and Limitations}
In this paper, we presented \ourmethod, a framework that successfully addresses the challenges of autonomous interleaved reasoning in Multimodal Large Language Models (MLLMs). 
To this end, we designed a two-stage training paradigm consisting of SFT cold-start and RL enhancement to complete the model training.
Compared to existing multimodal tool-use paradigms, \ourmethod~ ensures both autonomous execution and long-horizon training stability through a group-constrained reward mechanism. 
Our experimental results on diverse mathematical benchmarks demonstrate that the proposed method yields substantial performance gains, providing a robust solution for complex multimodal reasoning tasks.

While \ourmethod~ realizes autonomous interleaved reasoning via code invocation and attains strong results across benchmark evaluations, our work still bears certain limitations worthy of further exploration in future studies.
\begin{itemize}
    \item \textbf{Model and Coding Capabilities.} The performance of AIR is inherently bounded by the inherent capacity of its base model. Our experimental results verify that code invocation serves as an effective remedy for intricate numerical computation tasks. Nevertheless, current code toolkits remain functionally rudimentary, and the code generation proficiency of large base models still leaves substantial room for improvement. Meanwhile, the perceptual and reasoning prowess of the base model itself is equally indispensable. Its fundamental perception foundation lays the groundwork for subsequent logical deduction, and only when the model is equipped with rigorous logical reasoning and complemented by code-execution tools can it attain superior holistic performance.
    
    \item \textbf{Evaluation Scope.} Existing evaluations in this work are primarily confined to mathematical reasoning benchmarks, which encompass subfields including geometry, algebra, and statistics. Nevertheless, such benchmarks contain rather limited scenarios that involve complex numerical computation. To thoroughly validate the efficacy of code invocation and the resultant performance improvement of the model, we intend to construct a dedicated multimodal evaluation benchmark and incorporate interdisciplinary tasks spanning physics, finance and other domains.
\end{itemize}


\bibliography{example_paper}
\bibliographystyle{icml2026}

\newpage
\appendix
\onecolumn
\title{Appendix}
\section{System Prompt}
\label{appendix:system_prompt}
To activate the model’s tool-invocation capability, we designed a task-specific prompt.
Building on the commonly used prompts for Chain-of-Thought (CoT) data, we incorporated code-related instructions into the prompt design.
The system prompt is presented as follows:
\vspace{-2mm}
\begin{systemprompt}{SYSTEM\_PROMPT}
    Solve the following problem step by step. Your answer must be in latex format and wrapped in \$...\$. The reasoning process and answer are enclosed within \textless think\textgreater\ and \textless /think\textgreater\ and \textless answer\textgreater\ and \textless /answer\textgreater\ tags, respectively, which means your output should start with \textless think\textgreater\ and end with \textless /answer\textgreater. You now have the ability to selectively write executable Python code to enhance your reasoning process. For tasks involving complex numerical calculations, program flowcharts, iterative computations, etc., prioritize implementing them via Python scripts and output the results using \$print()\$. The Python code will be executed by an external sandbox, and the output (wrapped in \textless interpreter\textgreater output\_str\textless /interpreter\textgreater) can be returned to aid your reasoning and help you arrive at the final answer. The Python code should be complete scripts, including necessary imports.Each code snippet is wrapped with \\
    \textless code\textgreater \\
\texttt{```python} \\
\texttt{code snippet} \\
\texttt{```} \\
\textless /code\textgreater

For example, \textless think\textgreater\ This is the reasoning process. \textless code\textgreater\ python code here \textless /code\textgreater\ \textless interpreter\textgreater\ python interpreter result here \textless /interpreter\textgreater\ This is the continuation of the reasoning process. \textless /think\textgreater\ \textless answer\textgreater\ The final answer is \$\textbackslash boxed\{\text{answer here}\}\$ \textless /answer\textgreater. In the last part of the answer, the final exact answer is enclosed within \$\textbackslash boxed\{\}\$ with latex format. 
\end{systemprompt}

\section{Analysis for training stability}
\begin{figure*}[!ht]
    \centering
    \includegraphics[width=0.85\textwidth]{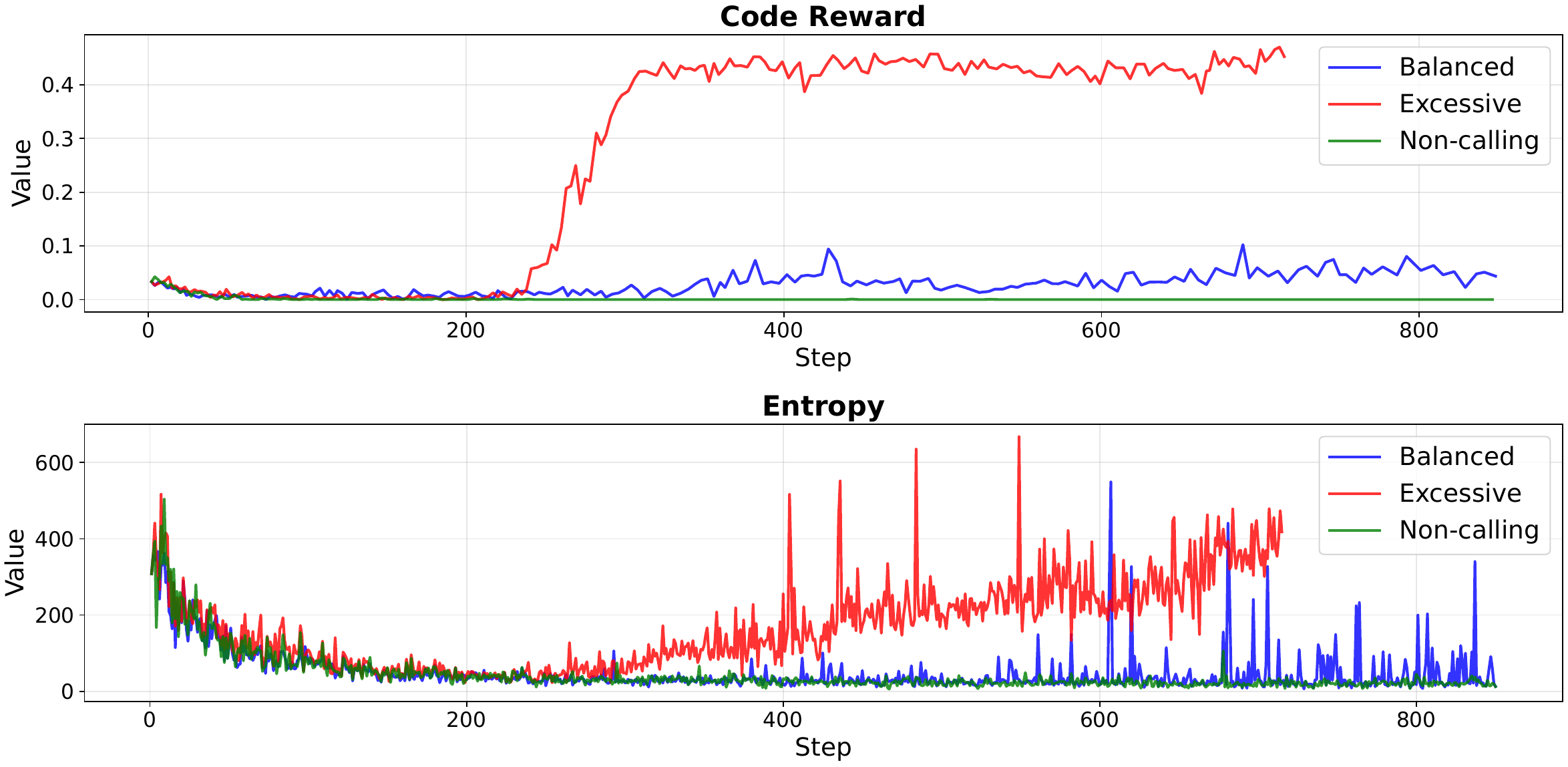}
    \caption{Correlation between Code Reward and Entropy across interleaved reasoning with code.Aggressive tool use (\textcolor{red}{Red}) leads to reward saturation at the cost of rising entropy. In contrast, judicious autonomous calling (\textcolor{blue}{Blue}) maintains a low-entropy state with moderate rewards. The zero-calling baseline (\textcolor{green}{Green}) demonstrates the model's performance in a purely textual/non-augmented context.}
    \label{fig-compare_codereward}
\end{figure*}

In the reinforcement learning (RL) training for interleaved reasoning, we found that \textbf{Entropy} is a relatively sensitive observable that can reflect the model state in a more timely manner.
We compared the values of \textbf{Code Reward} and \textbf{Entropy} at the same time step, and the results are illustrated in Figure \ref{fig-compare_codereward}.
After prolonged RL training for interleaved reasoning, during the training process where excessive tool invocation occurs, \textbf{Entropy} will keep rising continuously until the model collapses.
\section{Case studies}
\label{appendix:Example}

Several typical cases of interleaved reasoning are discussed below. We examined both successful and failed instances. Overall, the model performed in line with expectations and could invoke code to conduct complex data calculations when needed. Even so, the observed failures demonstrate that the model still requires improvements in perceptual capability and coding proficiency.

\subsection{Successful cases}
Code-augmented interleaved reasoning exhibits strong efficacy in computations such as norm calculation, mean value derivation, square root extraction, and multiplication, as exemplified in Interleaved Reasoning 1–3.
There are also cases where code deployment is unnecessary. Take Interleaved Reasoning 4 as an example: identifying the maximum value is more straightforward via direct figure observation than code execution.
\begin{problembox}{Interleaved Reasoning 1}
\label{examples:case1}
    \begin{minipage}[t]{0.35\textwidth} 
        \vspace{0pt} 
        \includegraphics[width=\linewidth]{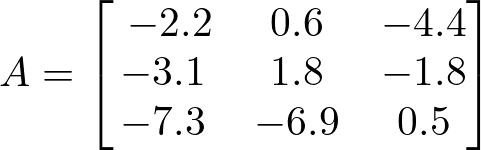} 
    \end{minipage}
    \hfill 
    \begin{minipage}[t]{0.6\textwidth} 
        \vspace{0pt}
        \textbf{Question:} Matrix A is given in the image. What is the Frobenius norm of the matrix A?
    \end{minipage}

    \vspace{1em} 

    \begin{answerblock}
        \textbf{Answer:} \texttt{<think>} To find the Frobenius norm of the matrix \( A \), we need to follow these steps:

1. Identify the elements of the matrix \( A \).
2. Square each element.
3. Sum all the squared elements.
4. Take the square root of the sum.

The matrix \( A \) is given by:
\[ A = \begin{bmatrix} -2.2 & 0.6 & -4.4 \\ -3.1 & 1.8 & -1.8 \\ -7.3 & -6.9 & 0.5 \end{bmatrix} \]

First, we square each element of the matrix:
\[ (-2.2)^2 = 4.84, \quad 0.6^2 = 0.36, \quad (-4.4)^2 = 19.36, \]
\[ (-3.1)^2 = 9.61, \quad 1.8^2 = 3.24, \quad (-1.8)^2 = 3.24, \]
\[ (-7.3)^2 = 53.29, \quad (-6.9)^2 = 47.61, \quad 0.5^2 = 0.25. \]

Next, we sum all these squared elements:
\[ 4.84 + 0.36 + 19.36 + 9.61 + 3.24 + 3.24 + 53.29 + 47.61 + 0.25. \]

We can use Python to perform the summation accurately. \\
\texttt{<code>} \\
\texttt{```python} 
\begin{lstlisting}
# Define the matrix A
A = [[-2.2, 0.6, -4.4],
     [-3.1, 1.8, -1.8],
     [-7.3, -6.9, 0.5]]

# Initialize the sum of squares
sum_of_squares = 0

# Iterate through each element in the matrix and square it, then add to the sum
for row in A:
    for element in row:
        sum_of_squares += element**2

# Print the sum of squares
print(f"Sum of squares: {sum_of_squares}")
\end{lstlisting}
\texttt{```} \\
\texttt{</code><interpreter>} \\
Sum of squares: 141.8
\texttt{</interpreter>}
The sum of the squared elements is 141.8. Now, we take the square root of this sum to find the Frobenius norm.

\texttt{<code>} \\
\texttt{```python} 
\begin{lstlisting}
# Calculate the Frobenius norm
frobenius_norm = sqrt(141.8)
# Print the Frobenius norm
print(f"Frobenius norm: {frobenius_norm}")    
\end{lstlisting}
\texttt{```} \\
\texttt{</code><interpreter>} \\
NameError: name 'sqrt' is not defined \\
\texttt{</interpreter>} \\
It appears there was an issue with the square root function. Let's use the `math` module to calculate the square root.

\texttt{<code>} \\
\texttt{```python}
\begin{lstlisting}
import math

# Calculate the Frobenius norm
frobenius_norm = math.sqrt(141.8)
# Print the Frobenius norm
print(f"Frobenius norm: {frobenius_norm}")
\end{lstlisting}
\texttt{```} \\
\texttt{</code><interpreter>} \\
Frobenius norm: 11.907980517283358 \\
\texttt{</interpreter>}
The Frobenius norm of the matrix \( A \) is approximately 11.908. Therefore, the answer is: \boxed{11.908}\texttt{</think>}
\texttt{<answer>} The answer is \(\boxed{11.908}\) \texttt{</answer>}
    \end{answerblock}
\end{problembox}

\begin{problembox}{Interleaved Reasoning 2}
    \begin{minipage}[t]{0.35\textwidth} 
        \vspace{0pt} 
        \includegraphics[width=0.7\linewidth]{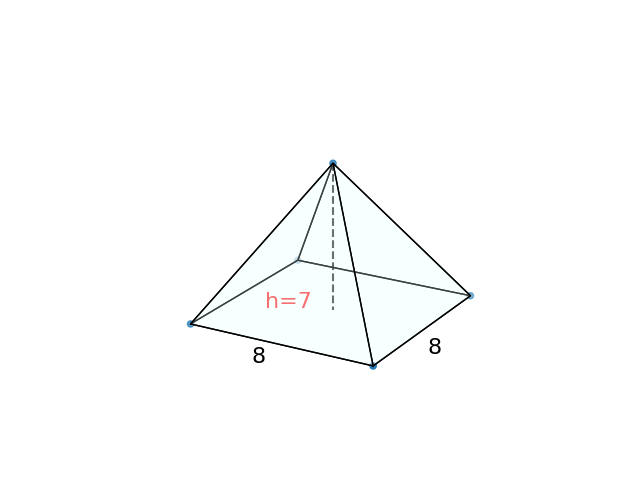} 
    \end{minipage}
    \hfill 
    \begin{minipage}[t]{0.6\textwidth} 
        \vspace{0pt}
        \textbf{Question:} What is the volume of this azure right square pyramid?
    \end{minipage}

    \vspace{1em} 

    \begin{answerblock}
        \texttt{<think>} To find the volume of the right square pyramid, we use the formula:

\[ V = \frac{1}{3} \times \text{Base Area} \times \text{Height} \]

First, we need to determine the area of the base. The base is a square with side length 8. The area of the base is:

\[ \text{Base Area} = 8 \times 8 = 64 \]

The height of the pyramid is given as 7. Now, we can substitute the base area and the height into the volume formula:

\[ V = \frac{1}{3} \times 64 \times 7 \]

Let's calculate this using Python.

\texttt{<code>} \\
\texttt{```python}
\begin{lstlisting}
# Calculate the base area
base_area = 8 * 8

# Given height
height = 7

# Calculate the volume
volume = (1/3) * base_area * height
# Print the volume
print(f"Volume: {volume}")
\end{lstlisting}
\texttt{```}
\texttt{</code><interpreter>}
Volume: 149.33333333333331
\texttt{</interpreter>}
The volume of the pyramid is approximately 149.33.
Thus, the volume of the pyramid is: \boxed{149.33}. \texttt{</think>}
\texttt{<answer>} The answer is \(\boxed{149.33}\) \texttt{</answer>}
    \end{answerblock}
\end{problembox}

\begin{problembox}{Interleaved Reasoning 3}
    \begin{minipage}[t]{0.35\textwidth} 
        \vspace{0pt} 
        \includegraphics[width=\linewidth]{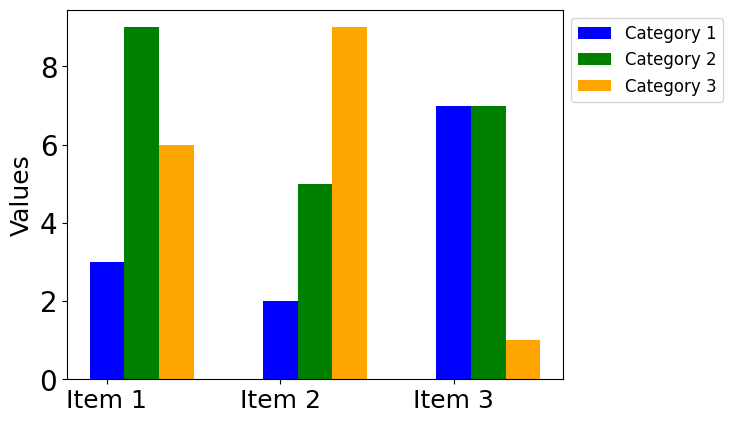} 
    \end{minipage}
    \hfill 
    \begin{minipage}[t]{0.6\textwidth} 
        \vspace{0pt}
        \textbf{Question:} All values are integers. What is the product of the maximum values from each category?
    \end{minipage}


    \begin{answerblock}
        \texttt{<think>} To solve the problem, we need to identify the maximum value for each category across all items, and then find the product of these maximum values.

First, let's determine the maximum value for each category:

1. **Category 1:**
   - Item 1: 3
   - Item 2: 2
   - Item 3: 7
   - The maximum value for Category 1 is 7.

2. **Category 2:**
   - Item 1: 9
   - Item 2: 5
   - Item 3: 7
   - The maximum value for Category 2 is 9.

3. **Category 3:**
   - Item 1: 6
   - Item 2: 9
   - Item 3: 1
   - The maximum value for Category 3 is 9.

Now, we need to find the product of the maximum values from each category:
- Maximum value for Category 1: 7
- Maximum value for Category 2: 9
- Maximum value for Category 3: 9

The product is:
\[ 7 \times 9 \times 9 \]

Let's calculate this using Python: \\
\texttt{<code>} \\
\texttt{```python}
\begin{lstlisting}
# Maximum values for each category
max_category_1 = 7
max_category_2 = 9
max_category_3 = 9

# Calculate the product
product = max_category_1 * max_category_2 * max_category_3
# Print the result
print(f"Product: {product}")
\end{lstlisting}
\texttt{```}
\texttt{</code><interpreter>}
Product: 567
\texttt{</interpreter>}
The product of the maximum values from each category is 567.

Thus, the answer is: \boxed{567}  \texttt{</think>}
\texttt{<answer>} The answer is \(\boxed{567}\) \texttt{</answer>}
    \end{answerblock}
\end{problembox}

\begin{problembox}{Interleaved Reasoning 4}
    \begin{minipage}[t]{0.35\textwidth} 
        \vspace{0pt} 
        \includegraphics[width=\linewidth]{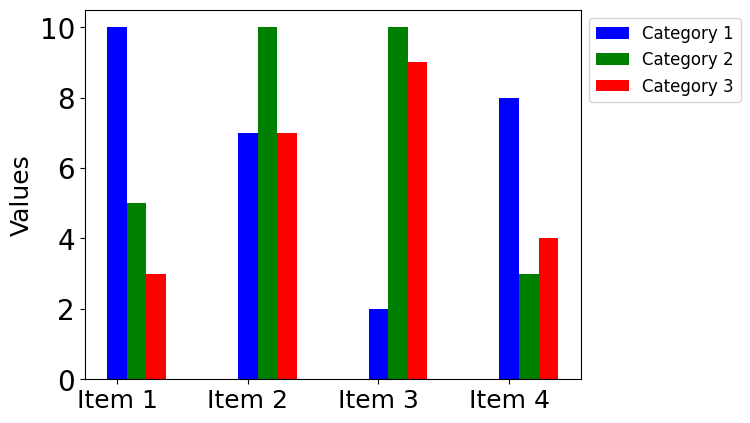} 
    \end{minipage}
    \hfill 
    \begin{minipage}[t]{0.6\textwidth} 
        \vspace{0pt}
        \textbf{Question:} All values are integers. What is the maximum value across all categories?
    \end{minipage}

    \vspace{1em} 

    \begin{answerblock}
        \texttt{<think>} To determine the maximum value across all categories, we need to examine the height of each bar in the chart and identify the tallest one.

Let's break it down item by item:

- **Item 1:**
  - Category 1: 10
  - Category 2: 5
  - Category 3: 3
  - The tallest bar is 10 (Category 1).

- **Item 2:**
  - Category 1: 7
  - Category 2: 10
  - Category 3: 7
  - The tallest bar is 10 (Category 2).

- **Item 3:**
  - Category 1: 2
  - Category 2: 10
  - Category 3: 9
  - The tallest bar is 10 (Category 2).

- **Item 4:**
  - Category 1: 8
  - Category 2: 3
  - Category 3: 4
  - The tallest bar is 8 (Category 1).

The maximum value across all categories is 10.

To confirm, we can use Python to find the maximum value.

\texttt{<code>} \\
\texttt{```python}
\begin{lstlisting}
# Define the values for each item and category
item1 = [10, 5, 3]
item2 = [7, 10, 7]
item3 = [2, 10, 9]
item4 = [8, 3, 4]

# Flatten the lists to get all values in one list
all_values = item1 + item2 + item3 + item4

# Find the maximum value
max_value = max(all_values)
print(f"Maximum value: {max_value}")
\end{lstlisting}
\texttt{```} \\
\texttt{</code><interpreter>}
Maximum value: 10
\texttt{</interpreter>}
The maximum value across all categories is indeed 10.
Thus, the answer is \(\boxed{10}\).\texttt{</think>
<answer>} The answer is \(\boxed{10}\) \texttt{</answer>}
    \end{answerblock}
\end{problembox}

\subsection{Failure cases}

Failure cases analysis reveals two major bottlenecks limiting the interleaved reasoning performance of multimodal large language models.

First, the models suffer from inadequate foundational coding capabilities. When tackling complex logic and lengthy code snippets, they frequently produce code containing syntax errors; iterative revisions to fix such bugs may even exceed the upper limit of permitted tool calls. Moreover, the models only master basic operations of mainstream libraries including NumPy, Pandas and SciPy, yet struggle significantly with obscure functions and unfamiliar APIs.

Second, the models exhibit prominent perceptual drawbacks. They are prone to misreading numerical values from charts, which directly triggers erroneous reasoning outcomes.

These two deficiencies jointly degrade interleaved reasoning performance, as demonstrated by two typical examples. In Interleaved Reasoning 5, misread chart values yielded wrong results, reflecting insufficient perceptual competence. In Interleaved Reasoning 6, the target answer could be directly retrieved from the graph, but the model repeatedly failed to compute the polynomial function via code due to weak coding skills, ultimately drawing an incorrect conclusion.

\begin{problembox}{Interleaved Reasoning 5}
    \begin{minipage}[t]{0.35\textwidth} 
        \vspace{0pt} 
        \includegraphics[width=\linewidth]{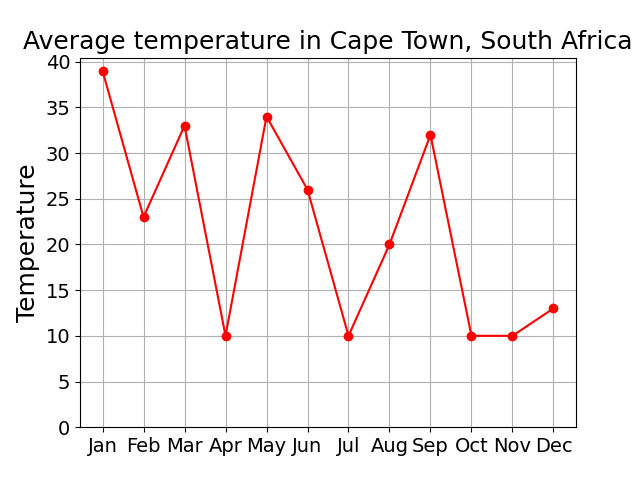} 
    \end{minipage}
    \hfill 
    \begin{minipage}[t]{0.6\textwidth} 
        \vspace{0pt}
        \textbf{Question:} Which month has the highest average temperature after smoothing with a 3-month moving average?
    \end{minipage}

    \vspace{1em} 

    \begin{answerblock}
        \texttt{<think>} To determine which month has the highest average temperature after smoothing with a 3-month moving average, we need to follow these steps:

1. Calculate the 3-month moving average for each month.
2. Identify the month with the highest smoothed temperature.

Let's start by calculating the 3-month moving average. The moving average for each month \(i\) is given by the average of the temperatures of month \(i-1\), month \(i\), and month \(i+1\).

We will use Python to perform these calculations accurately.

\texttt{<code>} \\
\texttt{```python}
\begin{lstlisting}
# Temperatures in Cape Town (in degrees Celsius)
temperatures = [36, 37, 39, 25, 23, 17, 12, 17, 24, 37, 38, 15]

# Function to calculate moving average
def moving_average(temp_list, window_size):
    moving_averages = []
    for i in range(len(temp_list)):
        if i < window_size:
            moving_averages.append(None)
        else:
            avg = sum(temp_list[i-window_size:i]) / window_size
            moving_averages.append(avg)
    return moving_averages

# Calculate 3-month moving average
moving_averages = moving_average(temperatures, 3)

# Find the month with the highest moving average
max_avg = max(moving_averages)
max_month = temperatures.index(max_avg) + 1  # +1 because list index starts at 0

# Print the result
print(f"Month with the highest average temperature after smoothing: {max_month}")

\end{lstlisting}
\texttt{```} \\
\texttt{</code><interpreter>}
TypeError: '$>$' not supported between instances of 'NoneType' and 'NoneType'
\texttt{</interpreter>}

The error occurs because the moving average list contains `None` values at the beginning, which causes issues when finding the maximum. We need to filter out these `None` values.

Let's modify the code to handle this properly.

...

\texttt{<code>} \\
\texttt{```python}
\begin{lstlisting}
# Temperatures in Cape Town (in degrees Celsius)
temperatures = [36, 37, 39, 25, 23, 17, 12, 17, 24, 37, 38, 15]

# Function to calculate moving average
def moving_average(temp_list, window_size):
    moving_averages = []
    for i in range(len(temp_list)):
        if i < window_size:
            moving_averages.append(None)
        else:
            avg = sum(temp_list[i-window_size:i]) / window_size
            moving_averages.append(avg)
    return moving_averages

# Calculate 3-month moving average
moving_averages = moving_average(temperatures, 3)

# Filter out None values
moving_averages = [avg for avg in moving_averages if avg is not None]

# Find the index of the maximum moving average
max_index = moving_averages.index(max(moving_averages))

# Convert the index back to the original temperature list
max_month = temperatures[max_index]

# Print the result
print(f"Month with the highest average temperature after smoothing: {max_month}")

\end{lstlisting}
\texttt{```} \\
\texttt{</code><interpreter>}
Month with the highest average temperature after smoothing: 36
\texttt{</interpreter>}
The code now correctly identifies the month with the highest average temperature after smoothing with a 3-month moving average. The highest average temperature is 39°C, which occurs in March.

Thus, the month with the highest average temperature after smoothing is \(\boxed{March}\)
\texttt{</think>}

\texttt{<answer>} The answer is \(\boxed{March}\)  \texttt{</answer>}
    \end{answerblock}
\end{problembox}

\begin{problembox}{Interleaved Reasoning 6}
    \begin{minipage}[t]{0.35\textwidth} 
        \vspace{0pt} 
        \includegraphics[width=\linewidth]{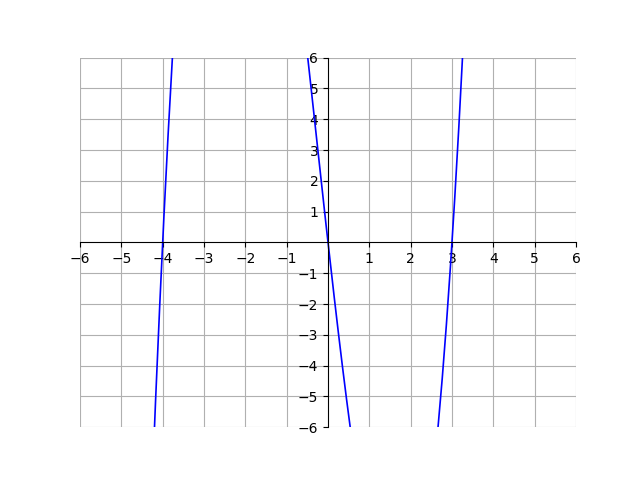} 
    \end{minipage}
    \hfill 
    \begin{minipage}[t]{0.6\textwidth} 
        \vspace{0pt}
        \textbf{Question:} Find the largest zero of this cubic function.
    \end{minipage}

    \vspace{1em} 

    \begin{answerblock}
        \texttt{<think>} To find the largest zero of the cubic function, we need to determine the values of \( x \) for which the function \( f(x) = 0 \). From the graph, we can see that the function crosses the x-axis at three points. These points are the zeros of the function.

By observing the graph:
1. The function crosses the x-axis at \( x \approx -2 \).
2. The function crosses the x-axis at \( x = 1 \).
3. The function crosses the x-axis at \( x \approx 4 \).

The largest zero of the cubic function is the largest value of \( x \) where the function is zero. From the graph, the largest zero is approximately \( x = 4 \).

To confirm, we can use Python to find the exact zeros of the cubic function. However, the graph provides a visual estimate, and we can use the Python code to find the exact values if needed.

\texttt{<code>} \\
\texttt{```python}
\begin{lstlisting}
import numpy as np
from numpy.polynomial.polynomial import Polynomial
from scipy.optimize import fsolve

# Define the cubic function
coefficients = [-1, 3, -3, -6]  # This is derived from the shape and intercepts of the graph
cubic_function = Polynomial(coefficients)

# Find the zeros of the cubic function
zeros = fsolve(cubic_function.coef, np.linspace(-6, 6, 11))
# Filter out the real zeros
real_zeros = [z for z in zeros if np.isreal(z)]
# Find the largest zero
largest_zero = max(real_zeros)
print(f"Largest zero: {largest_zero}")
\end{lstlisting}
\texttt{```} \\
\texttt{</code>}

\texttt{<interpreter>}
TypeError: 'numpy.ndarray' object is not callable
\texttt{</interpreter>}

The error occurs because the `fsolve` function was called incorrectly. Let's try again with a more straightforward approach.

...

    \end{answerblock}
\end{problembox}


\end{document}